# Inducing Probabilistic Grammars by Bayesian Model Merging

Andreas Stolcke
Stephen Omohundro
International Computer Science Institute
1947 Center St., Suite 600
Berkeley, CA 94707
E-mail: {stolcke,om}@icsi.berkeley.edu

**Abstract**

We describe a framework for inducing probabilistic grammars from corpora of positive samples. First, samples are *incorporated* by adding ad-hoc rules to a working grammar; subsequently, elements of the model (such as states or nonterminals) are *merged* to achieve generalization and a more compact representation. The choice of what to merge and when to stop is governed by the Bayesian posterior probability of the grammar given the data, which formalizes a trade-off between a close fit to the data and a default preference for simpler models ('Occam's Razor'). The general scheme is illustrated using three types of probabilistic grammars: Hidden Markov models, class-based $n$-grams, and stochastic context-free grammars.

## 1 Introduction

Probabilistic modeling has become increasingly important for applications such as speech recognition, information retrieval, machine translation, and biological sequence processing. The types of models used vary widely, ranging from simple $n$-grams to Hidden Markov Models (HMMs) and stochastic context-free grammars (SCFGs). A central problem for these applications is to find suitable models from a corpus of samples.

Most common probabilistic models can be characterized by two parts: a discrete structure (e.g., the topology of an HMM, the context-free backbone of a SCFG), and a set of continuous parameters which determine the probabilities for the words, sentences, etc. described by the grammar. Given the discrete structure, the continuous parameters can usually be fit using standard methods, such as likelihood maximization. In the case of models with hidden variables (HMMs, SCFGs) estimation typically involves expectation maximization (EM) (Baum *et al.* 1970; Dempster *et al.* 1977; Baker 1979).

In this paper we address the more difficult first half of the problem: finding the discrete structure of a probabilistic model from training data. This task includes the problems of finding the topology of an HMM, and finding the set of context-free productions for an SCFG. Our approach is called *Bayesian model merging* because it performs successive merging operations on the substructures of a model in an attempt to maximize the Bayesian posterior probability of the overall model structure, given the data.

In this paper, we give an introduction to Bayesian model merging for probabilistic grammar inference, and demonstrate the approach on various model types. We also report briefly on some of the applications of the resulting learning algorithms primarily in the area of natural language modeling.

## 2  Bayesian Model Merging

*Model merging* (Omohundro 1992) has been proposed as an efficient, robust, and cognitively plausible method for building probabilistic models in a variety of cognitive domains (e.g., vision). The method can be characterized as follows:

- **Data incorporation**: Given a body of data $X$, build an initial model $M_0$ by explicitly accommodating each data point individually such that $M_0$ maximizes the likelihood $P(X|M)$. The size of the initial model will thus grow with the amount of data, and will usually not exhibit significant generalization.

- **Structure merging**: Build a sequence of new models, obtaining $M_{i+1}$ from $M_i$ by applying a *generalization* or *merging* operator $m$ that coalesces substructures in $M_i$, $M_{i+1} = m(M_i), i = 0, 1, \ldots$

The merging operation is dependent on the type of model at hand (as will be illustrated below), but it generally has the property that data points previously 'explained' by separate model substructures come to be accounted for by a single, shared structure. The merging process thereby gradually moves from a simple, instance-based model toward one that expresses structural generalizations about the data.

To guide the search for suitable merging operations we need a criterion that trades off the goodness of fit of data $X$ against the desire for 'simpler,' and therefore more general models. As a formalization of this tradeoff, we use the *posterior probability* $P(M|X)$ of the model given the data. According to Bayes' rule,

$$P(M|X) = \frac{P(M)P(X|M)}{P(X)} \quad ,$$

the posterior is proportional to the product of a *prior probability* term $P(M)$ and a *likelihood* term $P(X|M)$ (the denominator $P(X)$ does not depend on $M$ and can therefore be ignored for the purpose of maximizing). The likelihood is defined by the model semantics, whereas the prior has to be chosen to express the bias, or prior expectation, as to what the likely models are. This choice is domain-dependent and will be elaborated below.

Finally, we need a search strategy to find models with high (maximal, if possible) posterior probability. A simple approach here is

- **Best-first search:** Starting with the initial model (which maximizes the likelihood, but usually has a very low prior probability), explore all possible merging steps, and successively choose the one (greedy search) or ones (beam search) that give the greatest immediate increase in posterior. Stop merging when no further increase is possible (after looking ahead a few steps to avoid simple local maxima).

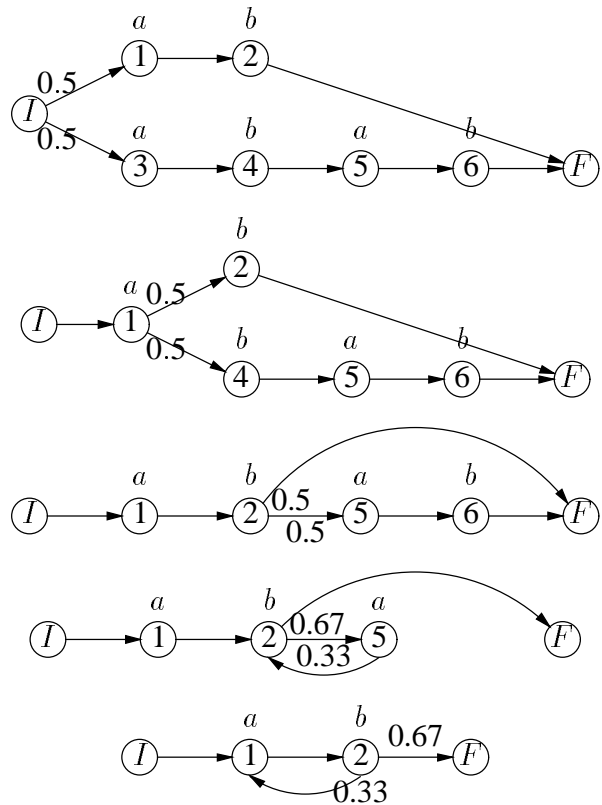

Figure 1: Model merging for HMMs.

In practice, to keep the working models of manageable size, we can use an *on-line* version of the merging algorithm, in which the data incorporation and the merging/search stages are interleaved.

We now make these concepts concrete for various types of probabilistic grammar.

# 3 Model merging applied to probabilistic grammars

## 3.1 Hidden Markov Models

Hidden Markov Models (HMMs) are a probabilistic form of non-deterministic finite-state models (Rabiner & Juang 1986). They allow a particularly straightforward version of the model merging approach.

**Data incorporation.** For each observed sample create a unique path between the initial and final states by assigning a new state to each symbol token in the sample. For example, given the data $X = \{ab, abab\}$, the initial model $M_0$ is shown at the top of Figure 1.

**Merging.** In a single merging step, two old HMM states are replaced by a single new state, which inherits the union of the transitions and emissions from the old states. Figure 1 shows four successive merges (where each new state is given the smaller of the two indices of its predecessors). The second, third and fifth models in the example have a smaller model structure without changing the generated probability distributions, whereas the fourth model effectively generalizes from the finite sample set to the infinite set $\{(ab)^n, n > 0\}$. The crucial point is that each of these models can be found by locally maximizing the posterior probability of the HMM, under a wide range of priors (see below). Also, further merging in the last model structure shown produces a large penalty in the likelihood term, thereby *decreasing* the posterior. The algorithm thus stops at this point.

**Prior distributions.** Our approach has been to choose relatively *uninformative* priors, which spread the prior probability across all possible HMMs without giving explicit preference to particular topologies. A model $M$ is defined by its *structure* (topology) $M_S$ and its continuous parameter settings $\theta_M$. The prior may therefore be decomposed as

$$P(M) = P(M_S)P(\theta_M|M_S) \quad .$$

Model structures receive prior probability according to their *description length*, i.e.,

$$P(M_S) \propto \exp(-\ell(M_S)),$$

where $\ell(M_S)$ is the number of bits required to encode $M_S$, e.g., by listing all transitions and emissions. The prior probabilities for $\theta_M$, on the other hand, are assigned using a Dirichlet distribution for each of the transition and emission multinomial parameters, similar to the Bayesian decision tree induction method of Buntine (1992). (The parameter prior effectively spreads the posterior probability as if a certain number of evenly distributed 'virtual' samples had been observed for each transition and emission.) For convenience we assume that the parameters associated with each state are *a priori* independent.

There are three intuitive ways of understanding why simple priors like the ones used here lead to higher posterior probabilities for simpler HMMs, other things being equal:

- Smaller topologies have a smaller description length, and hence a higher prior probability. This corresponds to the intuition that a larger structure needs to be 'picked' from among a larger range of possible equally sized alternatives, thus making each individual choice less probable *a priori*.

- Larger models have more parameters, thus making each particular parameter setting less likely (this is the 'Occam factor' (Gull 1988)).

- After two states have been merged, the effective amount of data per parameter increases (the evidence for the merged substructures is pooled). This shifts and peaks the posterior distributions for those parameters closer to their maximum likelihood settings.

These principles also apply *mutatis mutandis* to the other applications of model merging inference.

**Posterior computation.** Recall that the target of the inference procedure is the model *structure*, hence the goal is to maximize the posterior

$$P(M_S|X) \propto P(M_S)P(X|M_S)$$

The mathematical reason why one wants to maximize $P(M_S|X)$, rather than simply $P(M|X)$, is that for inference purposes a model with high posterior *structure* represents a better approximation to the Bayes-optimal procedure of averaging over all possible models $M$, including both structures and parameter settings (see Stolcke & Omohundro (1994:17f.) for details).

The evaluation of the second term above involves the integral over the parameter prior,

$$P(X|M_S) = \int_{\theta_M} P(\theta_M|M_S)P(X|M_S, \theta_M)d\theta_M,$$

which can be approximated using the common Viterbi assumption about sample probabilities in HMMs (which in our case tends to be correct due to the way the HMM structures are initially constructed).

**Applications and results.** We compared the model merging strategy applied to HMMs against the standard Baum-Welch procedure when applied to a fully parameterized, randomly initialized HMM structure. The latter represents one potential approach to the structure finding problem, effectively turning it into a parameter estimation problem, but it faces the problem of local maxima in the parameter space. Also, in a Baum-Welch approach the number of states in the HMM has to be known, guessed or estimated in advance, whereas the merging approach chooses that number adaptively from the data. Both approaches need to be evaluated empirically.

First, we tested the two methods on a few simple regular languages that we turned into HMMs by assigning uniform probabilities to their corresponding finite-state automata. Training proceeded using either random or 'structure covering' sets of samples. The merging approach reliably inferred these admittedly simple HMM structures. However, the Baum-Welch estimator turned out to be extremely sensitive to the initial parameter settings and failed on more than half of the trials to find a reasonable structure, both with minimal and redundant numbers of states.[1]

Second, we tested merging and Baum-Welch (and a number of other methods) on a set of naturally occurring data that might be modeled by HMMs. The task was to derive phonetic pronunciation models from available transcriptions in the TIMIT speech database. In this case, the Baum-Welch-derived model structures turned out to be close in generalization performance to the slightly better merged models (as measured by cross-entropy on a test set).[2] However, to achieve this performance the Baum-Welch HMMs made use of about twice as many transitions as the more compact merged HMMs, which would have a serious impact on potential applications of such models in speech recognition.

Finally, the HMM merging algorithm was integrated into the training of a medium-scale spoken language understanding system (Wooters & Stolcke 1994). Here, the algorithm also serves the purpose of inducing multi-pronunciation word models from speech data, but it is now coupled with a separate process that estimates the acoustic emission likelihoods for the HMM states. The goal of this setup was to improve the system's performance over a comparable system that used only the standard single-pronunciation HMMs for each word, while remaining practical in terms of training cost and recognition speed. By using the more complex, merged HMMs the word error was indeed reduced significantly (from 40.6% to 32.1%), indicating that the pronunciation models produced by the merging process were at least adequate for this kind of task.

---

[1] Case studies of the structures, under-generalizations and overgeneralizations found in this experiment can be found in Stolcke & Omohundro (1994).

[2] argued that this domain is slightly simpler, since it contains, for example, no looping HMM structures.

## 3.2 Class-based $n$-gram Models

Brown *et al.* (1992) describe a method for building class-based $n$-gram models from data. Such models express the transition probabilities between words not directly in terms of individual word types, but rather between word categories, or classes. Each class, in turn, has fixed 'emission' probabilities for the individual words. One potential advantage of this approach is that it can drastically reduce the number of parameters associated with ordinary $n$-gram models, by effectively sharing parameters between similarly distributed words.

To infer word classes automatically, Brown *et al.* (1992) suggest an algorithm that successively merges classes according to a maximum-likelihood criterion, until a target number of classes is reached. From our perspective we can cast their algorithm as an instance of model merging, the essential difference being the non-Bayesian (likelihood-based) criterion guiding the merging and stopping. In fact, in retrospect, class merging in $n$-gram grammars can be understood as a special case of HMM merging. A class-based $n$-gram model can be straightforwardly expressed as a special form of HMM in which each class corresponds to a state, and transition probabilities correspond to class $n$-gram probabilities.

## 3.3 Stochastic Context-Free Grammars

Based on the model merging approach to HMM induction, we have extended the algorithm to apply to stochastic context-free grammars (SCFGs), the probabilistic generalization of CFGs (Booth & Thompson 1973; Jelinek *et al.* 1992). A more detailed description of SCFG model merging can be found in Stolcke (1994).

**Data incorporation.** To incorporate a new sample string into a SCFG we can simply add a top-level production (for the start nonterminal $S$) that covers the sample precisely. For example, the grammar at the top of Figure 2 arises from the samples $\{ab, aabb, aaabbb\}$. Instead of letting terminal symbols appear in production right-hand sides, we also create one nonterminal for each observed terminal, which simplifies the merging operators.

**Merging.** The obvious analog of merging HMM states is the merging of nonterminals in a SCFG. This is indeed one of the strategies used to generalize a given SCFG, and it can potentially produce inductive 'leaps' by generating a grammar that generates more than its predecessor, while reducing the size of the grammar.

However the hallmark of context-free grammars are the hierarchical, center-embedding structures they can represent. We therefore introduce a second operator called *chunking*. It takes a given sequence of nonterminals and abbreviates it using a newly created nonterminal, as illustrated by the sequence $AB$ in the second grammar of Figure 2. In that example, one more chunking step, followed by two merging steps produces a grammar for the language

$$
\begin{aligned}
S &\rightarrow AB \\
&\rightarrow AABB \\
&\rightarrow AAABBB \\
A &\rightarrow a \\
B &\rightarrow b
\end{aligned}
$$

Chunk $(AB) \rightarrow X$:

$$
\begin{aligned}
S &\rightarrow X \\
&\rightarrow AXB \\
&\rightarrow AAXBB \\
X &\rightarrow AB
\end{aligned}
$$

Chunk $(AXB) \rightarrow Y$:

$$
\begin{aligned}
S &\rightarrow X \\
&\rightarrow Y \\
&\rightarrow AYB \\
X &\rightarrow AB \\
Y &\rightarrow AXB
\end{aligned}
$$

Merge $S, Y$:

$$
\begin{aligned}
S &\rightarrow X \\
&\rightarrow ASB \\
X &\rightarrow AB
\end{aligned}
$$

Merge $S, X$:

$$
\begin{aligned}
S &\rightarrow AB \\
&\rightarrow ASB
\end{aligned}
$$

Figure 2: Model merging for SCFGs.

$\{a^n b^n, n > 0\}$. (The probabilities in the grammar are implicit in the usage counts for each production, and are not shown in the figure.)

**Priors.** As before, we split the prior for a grammar $M$ into a contribution for the structural aspects $M_S$, and one for the continuous parameter settings $\theta_M$. The goal is to maximize the posterior of the structure given the data, $P(M_S|X)$. For $P(M_S)$ we again use a description length-induced distribution, obtained by a simple enumerative encoding of the grammar productions (each occurrence of a nonterminal contributes $\log N$ bits to the description length, where $N$ is the number of nonterminals). For $P(\theta_M|M_S)$ we observe that the production probabilities associated with a given left-hand side form a multinomial, and so we use symmetrical Dirichlet priors for these parameters.

| Language | Sample no. | Grammar | Search |
|---|---|---|---|
| Parentheses | 8 | $S \rightarrow () \mid (S) \mid SS$ | BF |
| $a^{2n}$ | 5 | $S \rightarrow aa \mid SS$ | BF |
| $(ab)^n$ | 5 | $S \rightarrow ab \mid SS$ | BF |
| $a^n b^n$ | 5 | $S \rightarrow ab \mid aSb$ | BF |
| $wcw^R, w \in \{a,b\}^*$ | 7 | $S \rightarrow c \mid aSa \mid bSb$ | BS(3) |
| Addition strings | 23 | $S \rightarrow a \mid b \mid (S) \mid S+S$ | BS(4) |
| Shape grammar | 11 | $S \rightarrow dY \mid bYS$ <br> $Y \rightarrow a \mid cY$ | BS(4) |
| Basic English | 25 | $S \rightarrow$ I am $A$ \| he $T$ \| she $T$ \| it $T$ <br> $\rightarrow$ they $V$ \| you $V$ \| we $V$ <br> $\rightarrow$ this $C$ \| that $C$ <br> $T \rightarrow$ is $A$ <br> $V \rightarrow$ are $A$ <br> $Z \rightarrow$ man \| woman <br> $A \rightarrow$ there \| here <br> $C \rightarrow$ is a $Z$ \| $ZT$ | BS(3) |

Table 1: Test grammars from Cook *et al.* (1976). Search methods are indicated by *BF* (best-first) or *BS(n)* (beam search with width $n$).

**Search.** In the case of HMMs, a greedy merging strategy (always pursuing only the locally most promising choice) seems to give generally good results. Unfortunately, this is no longer true in the extended SCFG merging algorithm. The chief reason is that chunking steps typically require several following merging steps and/or additional chunking steps to improve a grammar's posterior score. To account for this complication, we use a more elaborate *beam search* that considers a number of relatively good grammars in parallel, and stops only after a certain neighborhood of alternative models has been search without producing further improvements. The experiments reported below use small beam widths (between 3 and 10).

**Formal language experiments.** We start by examining the performance of the algorithm on example grammars found in the literature on other CFG induction methods. Cook *et al.* (1976) use a collection of techniques related to ours for inferring probabilistic CFGs from sample distributions, rather than absolute sample counts (see discussion in the next section). These languages and the inferred grammars are summarized in Table 1. They include classic textbook examples of CFGs (the parenthesis language, arithmetic expressions) as well as simple grammars meant to model empirical data.

We replicated Cook's results by applying the algorithm to the same small sets of high probability strings as used in Cook *et al.* (1976). (The number of distinct sample strings is given in the second column of Table 1.) Since the Bayesian framework makes use of the actual observed sample counts, we scaled these to sum to 50 for each training corpus.

The Bayesian merging procedure produced the target grammars in all cases, using different levels of sophistication in the search strategy (as indicated by column 4 in Table 1). Since Cook's algorithm uses a very different, non-Bayesian formalization of the data fit vs. grammar complexity trade-off we can conclude that the example grammars must be quite robust to a variety of 'reasonable' implementations of this trade-off.

A more difficult language that Cook *et al.* (1976) list as beyond the scope of their algorithm can also be inferred, using beam search: the palindromes $ww^R, w \in \{a, b\}^*$. We attribute this improvement to the more flexible search techniques used.

**Natural Language syntax.** An obvious question arising for SCFG induction algorithms is whether they are sufficient for deriving adequate models from realistic corpora of naturally occurring samples, i.e., to automatically build models for natural language processing applications. Preliminary experiments on such corpora have yielded mixed results, which lead us to conclude that additional methods will be necessary for success in this area. A fundamental problem is that available data will typically be *sparse* relative to the complexity of the target grammars, i.e., not all constructions will be represented with sufficient coverage to allow the induction of correct generalizations. We are currently investigating techniques to incorporate additional, independent sources of generalization. For example, a part-of-speech tagging phase prior to SCFG induction proper could reduce the work of the merging algorithm considerably.

Given these difficulties with large-scale natural language applications, we have resorted to smaller experiments that try to determine whether certain fundamental structures found in NL grammars can in principle be identified by the Bayesian framework proposed here. In Stolcke (1994) a number of phenomena are examined, including

**Lexical categorization** Nonterminal merging assigns terminal symbols to common nonterminals whenever there is substantial overlap in the contexts in which they occur.

**Phrase structure abstraction** Standard phrasal categories such as noun phrases, prepositional and verb phrases are created by chunking because they allow a more compact description of the grammar by abbreviating common collocations, and/or because they allow more succinct generalizations (in combination with merging) to be stated.

**Agreement** Co-variation in the forms of co-occurring syntactic or lexical elements (e.g., number agreement between subject and verbs in English) is induced by merging of nonterminals. However, even in this learning framework it becomes clear that CFGs (as opposed to, say, feature-base grammar formalisms) are an inadequate representation for these phenomena. The usual blow-up in grammar size to represent agreement in CFG form can also cause the wrong phrase structure bracketing to be prefered by the simplicity bias.

**Recursive phrase structure** Recursive and iterative productions for phenomena such as embedded relative clauses can be induced using the chunking and merging operators.

We conclude with a small grammar exhibiting recursive relative clause embedding, from Langley (1994). The target grammar has the form

```
S  --> NP VP
VP --> Verb NP
NP --> Art Noun
   --> Art Noun RC
```

```
    RC   --> Rel VP
    Verb --> saw | heard
    Noun --> cat | dog | mouse
    Art  --> a | the
    Rel  --> that
```

with uniform probabilities on all productions.

Chunking and merging of 100 random samples produces a grammar that is weakly equivalent to the above grammar. It also produced essentially identical phrase structure, except for a more compact implementation of the recursion through `RC`:

```
    S   --> NP VP
    VP  --> V NP
    NP  --> DET N
        --> NP RC
    RC  --> REL VP
    DET --> a
        --> the
    N   --> cat
        --> dog
        --> mouse
    REL --> that
    V   --> heard
        --> saw
```

## 4 Related work

Many of the ingredients of the model merging approach have been used separately in a variety of settings.

Successive merging of states (or state equivalence class construction) is a technique widely used in algorithms for finite-state automata (Hopcroft & Ullman 1979) and automata learning (Angluin & Smith 1983); a recent application to probabilistic finite-state automate is Carrasco & Oncina (1994).

Bell *et al.* (1990) and Ron *et al.* (1994) describe a method for learning deterministic finite-state models that is in a sense the opposite of the merging approach: successive state splitting. In this framework, each state represents a unique suffix of the input, and states are repeatedly refined by extending the suffixed represented, as long as this move improves the model likelihood by a certain minimum amount. The class of models thus learnable is restricted, since each state can make predictions based only on inputs within a bounded distance from the current input, but the approach has other advantages, e.g., the final number of states is typically smaller than for a merging algorithm, since the tendency is to overgeneralize, rather than undergeneralize. We are currently investigating state splitting as a complementary search operator in our merging algorithms.

Horning (1969) first proposed using a Bayesian formulation to capture the trade-off between grammar complexity and data fit. His algorithm, however, is based on searching for the grammar with the highest posterior probability by enumerating all possible grammars (such that one can tell after a finite number of steps when the optimal grammar has been found). Unfortunately, the enumeration approach proved to be infeasible for practical purposes.

The chunking operation used in SCFG induction is part of a number of algorithms aimed at CFG induction, including Cook *et al.* (1976), Wolff (1987), and Langley (1994), where it is typically paired with other operations that have effects similar to merging. However, only the algorithm of Cook *et al.* (1976) has probabilistic CFGs as the target of induction, and therefore merits a closer comparison to our approach.

A major conceptual difference of Cook's approach is that it is based on an information-theoretic quality measure that depends only on the *relative frequencies* of observed samples. The Bayesian approach, on the other hand, explicitly takes into account the *absolute frequencies* of the data. Thus, the *amount* of data available—not only its distribution—has an effect on the outcome. For example, having observed the samples $a, aa, aaa, aaaa$, a model of $\{a^n, n > 0\}$ is quite likely. On the other hand, if the same samples were observed a hundred times, with no other additional data, such a conclusion should be intuitively unlikely, although the sample strings themselves and their relative frequencies are unchanged. The Bayesian analysis confirms this intuition: a 100-fold sample frequency entails a 100-fold magnification of the log-likelihood loss incurred for any generalization, which would block the inductive leap to a model for $\{a^n, n > 0\}$.

Incidentally, one can use sample frequency as a principled device to control the degree of generalization in a Bayesian induction algorithm explicitly (Quinlan & Rivest 1989; Stolcke & Omohundro 1994).

## 5 Future directions

Since all algorithms presented here are of a generate-and-evaluate kind, they are trivial to integrate with external sources of constraints or information about possible candidate models. External structural constraints can be used to effectively set the prior (and therefore posterior) probability for certain models to zero. We hope to explore more informed priors and constraints to tackle larger problems, especially in the SCFG domain.

In retrospect, the merging operations used in our probabilistic grammar induction algorithms share a strong conceptual and formal similarity to those used by various induction methods for non-probabilistic grammars (Angluin & Smith 1983; Sakakibara 1990). Those algorithms are typically based on constructing equivalence classes of states based on some criterion of 'distinguishability.' Intuitively, the (difference in) posterior probability used to guide the Bayesian merging process represents a fuzzy, probabilistic version of such an equivalence criterion. This suggests looking for other non-probabilistic induction methods of this kind and adapting them to the Bayesian approach. A promising candidate we are currently investigating is the transducer inference algorithm of Oncina *et al.* (1993).

## 6 Conclusions

We have presented a Bayesian model merging framework for inducing probabilistic grammars from samples, by stepwise generalization from a sample-based ad-hoc model through successive merging operators. The framework is quite general and can therefore be instantiated for a variety of standard or novel classes of probabilistic models, as demonstrated here for HMMs and SCFGs.

The HMM merging variant, which is empirically more reliable for structure induction than Baum-Welch estimation, is being used successfully in speech modeling applications. The SCFG version of the model algorithm generalizes and simplifies a number of related algorithms that have been proposed previously, thus showing how the Bayesian posterior probability criterion can combine data fit and model simplicity in a uniform and principled way. The more complex model search space encountered with SCFGs also highlights the need for relatively sophisticated search strategies.